\def\year{2021}\relax
\documentclass[letterpaper]{article} 
\usepackage{aaai21}  
\usepackage{times}  
\usepackage{helvet} 
\usepackage{courier}  
\usepackage[hyphens]{url}  
\usepackage{graphicx} 
\usepackage{natbib}  
\usepackage{caption} 
\usepackage{booktabs}
\usepackage[caption=false]{subfig}
\usepackage{adjustbox}
\usepackage{amsmath,amssymb}
\newcommand{\fullname}{CARP\textsuperscript{e} Posterum}
\newcommand{\method}{CARP\textsuperscript{e}}

\usepackage[switch]{lineno}

\title{\fullname: A Convolutional Approach for Real-Time\\ Pedestrian Path Prediction}
\author {

        Mat\'{i}as Mendieta,
        Hamed Tabkhi \\
}
\affiliations {
    University of North Carolina at Charlotte\\
    mmendiet@uncc.edu, htabkhiv@uncc.edu 
}
\begin{document}

\maketitle

\begin{abstract}
Pedestrian path prediction is an essential topic in computer vision and video understanding. Having insight into the movement of pedestrians is crucial for ensuring safe operation in a variety of applications including autonomous vehicles, social robots, and environmental monitoring. Current works in this area utilize complex generative or recurrent methods to capture many possible futures. However, despite the inherent real-time nature of predicting future paths, little work has been done to explore accurate and computationally efficient approaches for this task. To this end, we propose a convolutional approach for real-time pedestrian path prediction, \method. It utilizes a variation of Graph Isomorphism Networks in combination with an agile convolutional neural network design to form a fast and accurate path prediction approach. Notable results in both inference speed and prediction accuracy are achieved, improving FPS considerably in comparison to current state-of-the-art methods while delivering competitive accuracy on well-known path prediction datasets.
\end{abstract}

\section*{Introduction}\label{intro}
Enabling algorithms with the ability to predict future trajectories of pedestrians has received increasing attention in recent years \cite{path_survey}. Such work is well warranted, with applications in societally impactful technologies like self-driving cars, social robotics, and environmental monitoring systems. However, this task has many challenging properties: 1) When choosing their future steps, pedestrians typically have an intrinsic goal from which they plan accordingly. Capturing this intent from outside observation requires a fundamental understanding of human movement. 2) Person-to-person social interactions often influence the future path of a pedestrian, for example, when avoiding collisions or traveling in groups. Therefore, modeling this social effect is imperative for robust path prediction. 3) Predicting the future is inherently time-sensitive, as the information is only useful for decision making if obtained quickly. Therefore, meeting real-time processing constraints is essential for the safety and usefulness of a path prediction algorithm.

Pioneering works have attempted to incorporate social effects in pedestrian trajectory prediction \cite{social_force,youll_never_walk_alone,who_are_you_with}. These approaches relied mainly on hand-crafted rules, and were often limited in scale and function. More recent works have focused on developing data-driven approaches to tackle the path prediction problem. Social LSTM \cite{slstm} formed a pooling mechanism with recurrent neural networks (RNNs) to provide social context to the prediction. Since then, many approaches have added the use of Generative Adversarial Networks (GANs) \cite{gan} within such frameworks, aiming to model the distribution of possible future trajectories \cite{sgan,idl,sophie}. Most recently, the work of Kosaraju et al. \cite{bigat} utilized a graph neural network to model the social situation in addition to an RNN-based GAN architecture. However, the existing approaches often have two major shortcomings. First, they rely on very complex models with many parameters, which makes the real-time execution on embedded devices nearly impossible. Second and more importantly, they use multiple runs over video frames that inherently violate the real-time nature of path prediction and limit applicability to real-world problems.


\begin{figure}[t]
\centering
    \begin{adjustbox}{max width=\columnwidth}
        \includegraphics{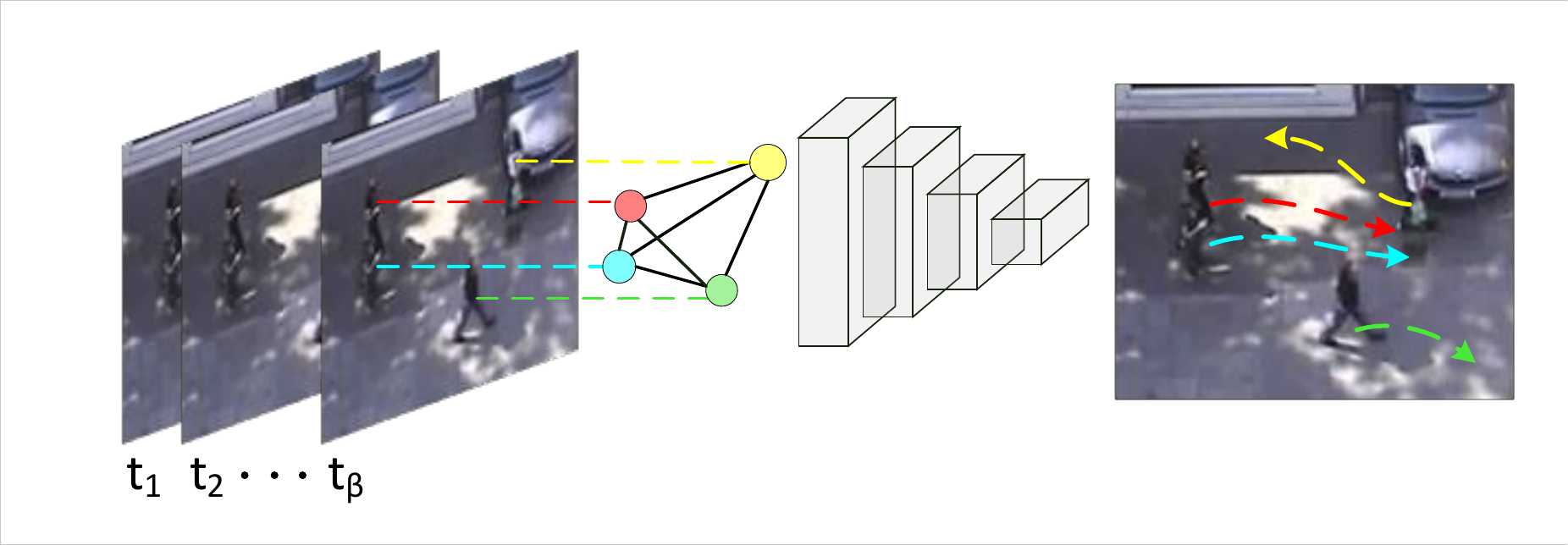}
    \end{adjustbox}
\caption{A high-level illustration of our proposed method, \method. Past pedestrian positions from $t_1$ to the current time step $t_\beta$ are fed into the model. This information is propagated through a graph and convolutional neural network in an end-to-end fashion, producing future predicted trajectories for the next $T$ time steps. In the depicted output, we show potential examples of intrinsic non-linearities (green), as well as social effects resulting from collision avoidance (yellow) and traveling groups (cyan and red).}
\label{fig:intro}
\end{figure}

To address the challenges of real-time path prediction, this paper proposes \fullname. \method~is a data-driven approach which effectively captures both intrinsic and social non-linearities of human trajectories, within real-time constraints. Figure \ref{fig:intro} shows the high-level mechanism of \method. Our method mainly consists of two networks, a graph neural network and a convolutional neural network, knitted together in an end-to-end fashion with efficiency in mind. \method~harnesses the strong and proven discriminative power of recently proposed Graph Isomorphism Networks \cite{gin} to gather social context, and an intentionally designed CNN architecture for effective path prediction. Our contributions are as follows:

\begin{itemize}
    \item A novel path prediction method, which captures both non-linear intrinsic and social effects.
    \item An agile end-to-end network architecture capable of real-time inference on low-power GPU and CPU devices.
\end{itemize}

Overall, notable results in both inference speed and prediction accuracy are achieved, improving FPS by at least 8x in comparison to current state-of-the-art methods while still producing competitive accuracy results on well-known path prediction datasets.

\section*{Related Work} \label{relatedOverall}

\subsection*{Pedestrian Path Prediction} \label{relatedPath}
Early works in pedestrian path prediction focused on the development of Gaussian processes and energy models to understand human behavior and movement \cite{social_force,youll_never_walk_alone,smooth_guassian,joint_modeling,who_are_you_with}. However, these methods often require many predefined rules for pedestrian interactions, and are limited to predicting a short time into the future. The classic work for activity forecasting \cite{irl_activity} approaches this task from a different angle. Instead of predicting specific locations of a trajectory, this method applies inverse control theory to determine the actions/motions of persons in a scene and form potential paths from such information. However, this method relies on static environments, and does not consider the dynamic social aspects of the environment.

Recurrent architectures are common options for recent works in path prediction, given their theoretical ability to capture an infinite history of inputs \cite{lstm,rnn_seq}. \cite{slstm,scene-lstm, srlstm, cidnn}, utilize Long Short-Term Memory (LSTM) based RNNs at the forefront of their approaches to understand and predict human trajectories. Liang et al. \cite{next} propose an LSTM-based joint trajectory and activity prediction system that incorporates scene segmentation maps, pedestrian visual features, and person keypoints to better inform both tasks. However, in practice, the infinite history capabilities of RNNs are largely absent and the forced sequential operation within the RNN limits its parallelization potential in modern hardware \cite{cnn2}. This challenges the effectiveness of RNNs versus purely convolutional networks in sequence modeling tasks, and makes such RNN approaches not ideal for deployable real-time inference.

Many recent works have focused on incorporating generative models for the path prediction problem. In \cite{sgan}, Gupta et al. built on the work of \cite{slstm} by integrating a social pooling mechanism into an LSTM-based GAN network. \cite{sophie} takes this work further, combining scene-level visual features with attention modules for physical and social relations. \cite{idl} aims specifically to capture the latent decision, or intrinsic elements, of pedestrian movements in a generative fashion with statistical sub-networks. In the graph domain, \cite{traj} forms a spatial-temporal graph architecture with a combination of Conditional Variational Autoencoders (CVAEs) \cite{cvae} as a generative model and LSTM units for the temporal dimension. Most recently, Kosaraju et al. form an RNN-based generative approach that includes the use of Bicycle-GANs \cite{bgan} and Graph Attention Networks (GAT) \cite{gat}. These generative approaches typically rely on the ability to repeatedly inference the model and generate many samples per pedestrian, which neglects the inherent real-time constrains of practical path prediction.


In contrast to these works, our method relies neither on recurrent nor generative architectures. We use a convolutional approach for hardware-friendliness and deterministic real-time inference. For social context, we formulate a Graph Neural Network (GNN) based on recent theoretical work in GNNs \cite{gin} to maximize its discriminative power. In this way, we tackle both intrinsic and social effects of pedestrian path prediction while achieving real-time inference capabilities on low-power devices.

\subsection*{Graph Neural Networks} \label{relatedGnn}
Graph Neural Networks (GNNs) have seen great progress in recent years. Naturally, GNNs aim to take advantage of the powerful learning ability of neural networks for non-Euclidean data. Data representations for molecular models or social media interactions are naturally inclined to graph representations, and therefore require a unique neural network definition. Typically, each node in the graph holds a feature, which is operated on across the graph structure. Such operations are utilized for a variety of objectives such as forming new node features, performing node classification, or completing graph-level classification \cite{survey}. 

GNNs are implemented with two major approaches, spectral and spatial \cite{survey}. Spectral methods, such as \cite{GCN,cheb}, utilize mathematical formulations rooted in graph signal processing to perform a graph Fourier Transform and subsequent convolution. However, these methods rely on the Laplacian eigenbasis, which is dependent on the graph structure. This property hinders the viability of spectral methods when dynamic graph structures are preferred.

Spatial methods. e.g. \cite{sage,gat,gin}, do not require such assumptions, as they operate on each node relative to its neighbors, allowing for dynamic graph structures. Typically, for a given node, the features of its neighbors are aggregated, and then combined with the current node feature. The aggregate and combine operations differ in the various spatial GNN formulations. This process can be repeated to form more abstract node representations, as well as increase the reach of a given node in a sparsely-connected graph. GraphSAGE \cite{sage} proposed an inductive learning framework with max-pooling aggregation across node features. Graph Attention Networks (GAT) \cite{gat} perform the aggregation and combine steps together using a weighted sum approach with attention. However, these GNN formulations have been largely based on empirical evidence, without a supporting theoretical foundation for optimal behavior. In their recent work \cite{gin}, Xu et al. investigate the theoretical properties of GNNs to determine optimal discriminative power based on the Weisfeiler-Lehman (WL) graph isomorphism test \cite{wl}. They present a new GNN formulation, termed Graph Isomorphism Networks (GIN), that are provably among the most expressive GNN variants. In this work, we use the GIN operators as a basis for our graph network and reformulate for use in path prediction.

\section*{Preliminary: Graph Isomorphism Networks} \label{preliminary}

For graph convolutional operations, the typical aim is to analyze a graph structure and the features of its nodes, producing meaningful representations in an embedding space across different graphs. Ideally, for a GNN to be maximally discriminative, two separate nodes should only map to the same location in the embedding space if all aspects of node and neighborhood are identical. These aspects include both the node features and neighborhood structure. Therefore, we expect to gather a unique feature embedding in all other cases. Testing for discriminative power in a GNN can be analogously drawn to the task of graph isomorphism tests, or distinguishing whether two graphs are topologically identical. 

A well-known test for determining such properties is the WL isomorphism test, as it has been found to effectively classify a broad class of graphs \cite{wl}. Therefore, Xu et al. \cite{gin} use the WL isomorphism test as a theoretical guide in determining the discriminative power of GNNs. In their work, the authors find that a GNN is as powerful as the WL isomorphism test if its aggregation (and graph level readout operation, as used in node classification) is injective. Therefore, the authors define a joint aggregation/combination operator as shown in Equation \ref{gineq}.

\begin{equation}\label{gineq}
    h_{i}^{\prime} = \phi \Bigg((1 + \epsilon) \cdot h_{i} + \sum_{j\in N(i)} h_j\Bigg)
\end{equation}

Here, node features $h_j$ are from nodes in the neighborhood $N(i)$. $h_i$ is the feature for node $i$, $\epsilon$ is a trainable parameter, $\phi$ indicates an Multilayer Perception (MLP), and $h_{i}^{\prime}$ is the updated node feature. In this work, we employ the findings of \cite{gin} and this joint aggregation/combination operator to formulate a graph for the pedestrian path prediction, as will be detailed in a later section.

\section*{\fullname: Method} \label{methodOverall}


The task of pedestrian path prediction is to predict the position of a pedestrian for $T$ time steps in the future given the past $\beta$ observed positions of the pedestrian. The goal is to accomplish this task as accurately as possible, while maintaining real-time inference capabilities. Two major factors that contribute to the future trajectory of a pedestrian are the intrinsic location goal of that pedestrian and the social context of the environment. We therefore aim to develop a model to capture these factors using the observed trajectories of all $P$ pedestrians in the scene at a given time step.

For the remainder of this paper, we will distinguish these various elements as follows: Past pedestrian trajectories take the form of absolute coordinates $A$ and relative coordinates $R$, defined as $A_{i}=\left\{\left(x_{i}^{t}, y_{i}^{t}\right) | t=1, \cdots, \beta\right\}$ and $R_{i}=\left\{\left(x_{i}^{t}-x_{i}^{1}, y_{i}^{t}-y_{i}^{1}\right) | t=1, \cdots, \beta\right\}$, $\forall i \in \{1, 2, \cdots, P\}$. The future trajectories $\hat{Y}$ of the pedestrians are predicted, and outputted as  $\hat{Y}_{i}=\left\{\left(x_{i}^{t}, y_{i}^{t}\right) | t=\beta+1, \cdots, T\right\}$, $\forall i \in \{1, 2, \cdots, P\}$. These predictions are compared with the ground truth future trajectories $Y$ for evaluation.

\subsection*{\fullname: Model Overview} \label{methodModelOverview}
\begin{figure*}[t]
\centering
    \begin{adjustbox}{max height=4in}
        \includegraphics{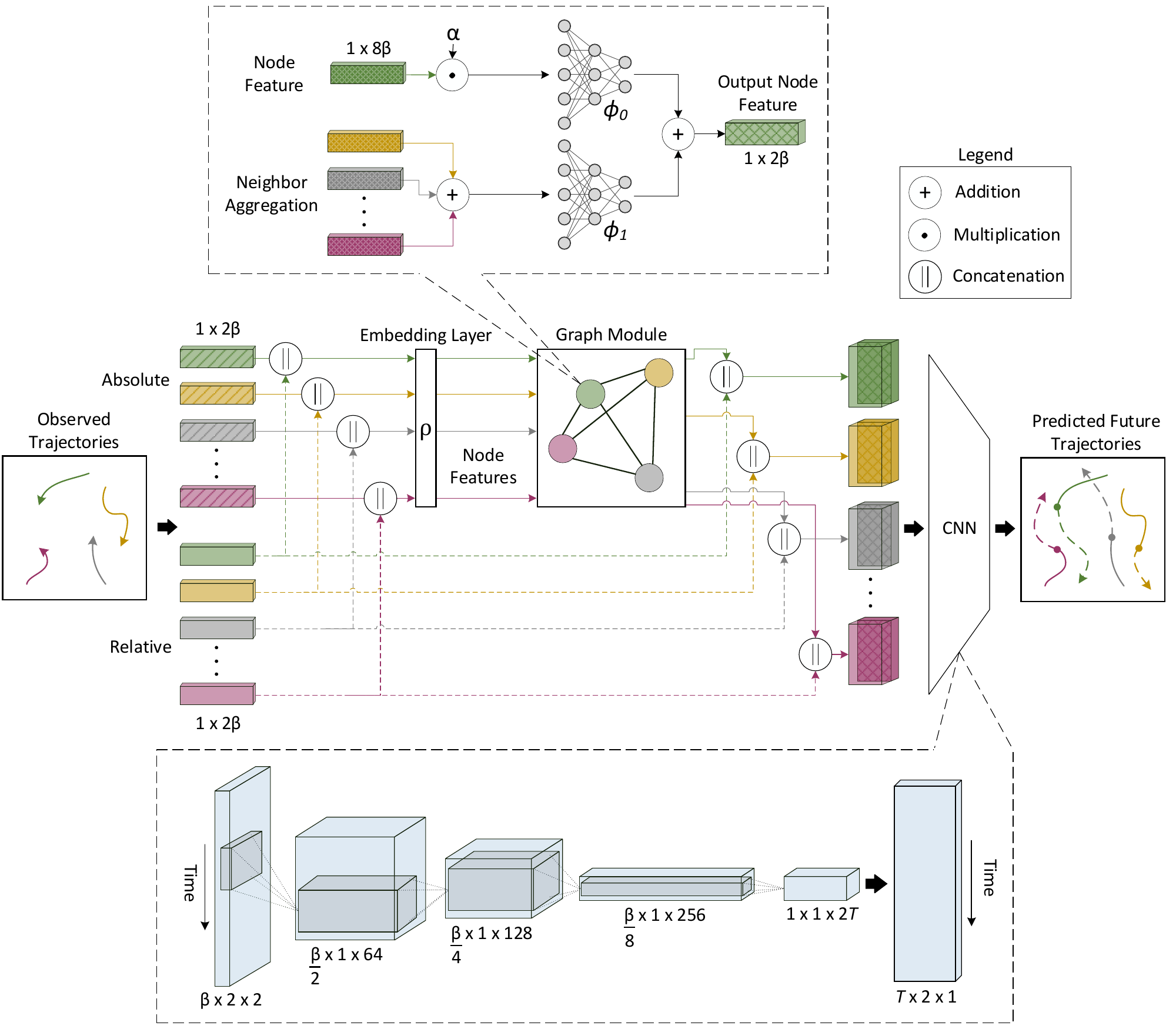}
    \end{adjustbox}
\caption{An overview of the full model architecture and data mechanisms for \method. Listed dimensions are in the form $row \times column \times channel$. Observed positions for $\beta$ time steps are formed into trajectories for each of the $P$ pedestrians present in the scene. Embedded absolute and relative trajectories are inputted into the graph module to gather social context features. The Prediction Module CNN then utilizes these features along with the relative coordinates of each pedestrian to produce informed future trajectory predictions.  In streaming applications, the algorithm receives detected coordinates as a stream of input, and maintains a buffer of the past and present coordinates for $\beta$ time steps. Therefore, in a sliding-window fashion over time, the algorithm predicts the future path for the next $T$ time steps.}
\label{fig:net}
\end{figure*}

Overall, \method's model consists of two main segments: 1) the Graph Module and 2) the Prediction Module. Figure \ref{fig:net} visualizes the full data mechanisms and module internals of \method. The role of the graph module is to produce features for each observed pedestrian that incorporate a broader social context across the scene. These features, along with the original observed trajectories of the pedestrians are both utilized by the prediction module to produce the future trajectories for each pedestrian. In this task, all trajectories for all pedestrians in the scene are inferred simultaneously, taking $P$ pedestrian features as input and outputting $P$ future trajectories in a single pass. We will explain each module and their functional details in the following sections.

\subsection*{\fullname: Graph Module} \label{sec:methodGraphModule}

\subsubsection{Graph Formulation.}
A graph $G = (V, E)$ is constructed, where $V$ and $E$ are the sets of nodes and edges respectively. All $P$ pedestrians in the scene are represented as nodes in $V = \{V_0, V_1, \cdots , V_P\}$. Each pedestrian in the graph has a corresponding node feature $h_i$ held within the graph structure. The GNN performs joint aggregation and combination operations on $G$ to produce an output set of node features $h^{\prime} = \{h_0^{\prime}, h_1^{\prime}, \cdots , h_P^{\prime}\}$. 

To form the input node feature, $A_i$ and $R_i$ are concatenated for a given pedestrian, and inferenced through a single fully-connected layer $\rho$. The absolute coordinates define the global position of the pedestrian, while the relative features act as a normalized form of input to better understand the pedestrian's past movement pattern.

\subsubsection{Graph Operation.}
Upon obtaining the node features, the graph is constructed as previously described. In order to maintain global context, the graph is fully connected. This allows the network to learn the relevant information needed, rather than predefining with handcrafted rules how relational connections should be made. To collect information across the graph, we define an aggregation and combination operation. The joint operation is represented in Equation \ref{gineqCustom}, based on the GIN operation described in the preliminary. In \cite{gin}, Xu et al. only employ one MLP in their base operation. However, we reason that abstracting the representation of the target node and the social context separately before combining will enable a deeper understanding and integration of neighboring nodes in context. Therefore, \method's graph operator performs an MLP operation on the summed neighborhood features and the node features with two separate MLPs $\phi_0$ and $\phi_1$. The MLP architectures are of two layers each for $\phi_0$ and $\phi_1$ in order to satisfy the universal approximation theorem \cite{uat1,uat2} and the recommendations for GIN operations as defined in \cite{gin}.

\begin{equation}\label{gineqCustom}
    h_{i}^{\prime} = \phi_0 \Big(\alpha \cdot h_{i}\Big) + \phi_1\bigg(\sum_{j\in N(i)} h_j\bigg)
\end{equation}

This process is illustrated in the top portion of Figure \ref{fig:net}, where $\alpha = 1 + \epsilon$ from Equation \ref{gineq}. Only a single graph operation is completed across the graph. This is done for two reasons. First, because the graph is fully connected, all pedestrians are accounted for in a single operation. Second, in aiming for real-time feasibility, limiting the number of operations allows our method to efficiently operate at scale. The output node features $h_{i}^{\prime}$  $\forall i \in \{1, 2, \cdots, P\}$ are subsequently employed in the Prediction Module.

\subsection*{\fullname: Prediction Module} \label{methodPredictionModule}
Typically, Recurrent Neural Networks (RNNs) are employed as the basis for state-of-the-art path prediction methods. Given the theoretical ability of RNNs to capture information along infinitely many time steps, such architectures have been frequently chosen for sequence-based problems, particularly the LSTM variant. However, recent works in the sequence modeling domain \cite{cnn2,cnn1, convseq} have found convolutional architectures to be advantageous over RNNs in many ways. Convolutional approaches to sequence modeling often form conceptually simpler networks, have more stable gradients and allow for greater parallelization, while producing comparable or improved accuracies on sequence data. Additionally, CNNs are well established for effectively capturing correlations in the spatial domain \cite{goodfellow}. The task of path prediction presents itself as both a spatially and temporally sensitive task, receiving and predicting $(x_{i}^{t}, y_{i}^{t})$ coordinate values across time. Therefore, we find that utilizing a convolutional architecture rather than an RNN may be more effective for the path prediction problem, while offering desirable hardware-friendly characteristics for real-time inference.

To this end, we form our Prediction Module with a simple CNN design to maximize spatial and temporal understanding, taking full advantage of the convolutional architecture approach. In our model, we first provide as input $S_i \in \mathbb{R}^{\beta \times 2\times 2}$, formed from the relative feature for a pedestrian $R_i \in \mathbb{R}^{2\beta}$ and their corresponding output node feature $h_i^{\prime} \in \mathbb{R}^{2\beta}$ concatenated together. In $S_i$, the $(x_{i}^{t}, y_{i}^{t})$ coordinate pairs are mapped with temporal order in the rows. We therefore map both the spatial coordinate information and temporal context into the 2D domain, where the CNN can advantageously correlate. This input structure provides the ability to naturally analyze the observed trajectory at various time granularites, adjusting filter size and stride, in a hardware-friendly fashion. In RNNs, such analysis would be impractical and computationally inefficient, requiring multiple LSTMs per pedestrian. To provide additional social context to the input of the Prediction Module, the output node feature $h_i^{\prime}$ is placed as the second channel in $S_i \in \mathbb{R}^{\beta \times 2\times 2}$.

The layers of the network are designed to capture changes in velocity and position with a bottom-up approach. First, a 2x2 filter is convolved across the input, as illustrated in Figure \ref{fig:net}. By convolving across just two time steps for each kernel, we emphasis model awareness of the high frequency movement and velocity changes over the observed period. As the feature progresses through the network, 2x1 filters are employed to find lower frequency trends, gathering the context of the trajectory across more time steps in the subsequent compressed representations. After the third network layer, \method~ produces a tensor in $\mathbb{R}^{\frac{\beta}{8} \times 1\times 2T}$ that obtains a holistic understanding of the observed trajectory. A subsequent $\frac{\beta}{8}$x1 convolution transforms this feature into the predicted trajectory.

\section*{Experiments} \label{experiments}

\subsection*{Evaluation Methodology}
We evaluate our model on two widely used datasets in the path prediction domain, ETH \cite{eth} and UCY \cite{ucy}. The ETH dataset is split into two portions (ETH, HOTEL), and UCY is split into three portions (UNIV, ZARA1, ZARA2). All portions are from distinct scenes other than ZARA1 and ZARA2, which are the same scene at different times. These datasets consist of a variety of pedestrian navigation situations, including many non-linear behaviors and social interactions. We utilize the same data and evaluation procedures as in \cite{sgan}, and commonly used in path prediction works \cite{slstm,bigat,next,sophie}. Therefore, a leave-one-out approach is applied for training and testing among the five scenarios. The data is collected as real-world coordinates in meters, with observations taken for 8 time steps (3.2 seconds) and predictions made for the next 12 time steps (4.8 seconds). Two metrics are utilized for quantitative evaluation on the ETH/UCY datasets:

\begin{itemize}
    \item Average Displacement Error (ADE) - The average L2 distance between the ground truth $(x, y)$ positions $Y$ and predicted $\hat{Y}$ for all $T$ predicted time steps over all $P$ pedestrians.
        \begin{equation}
          \mathrm{ADE}=\frac{\sum_{i=1}^{P} \sum_{t=1}^{T}\left\|Y_{i}^{t}-\hat{Y}_{i}^{t}\right\|_{2}}{P * T}  
        \end{equation}

    \item Final Displacement Error (FDE) - The average L2 distance between the ground truth $(x, y)$ positions $Y$ and predicted $\hat{Y}$ for only the final time step $T$ over all $P$ pedestrians.
        \begin{equation}
            \mathrm{FDE}=\frac{\sum_{i=1}^{P}\left\|Y_{i}^{T}-\hat{Y}_{i}^{T}\right\|_{2}}{P}
        \end{equation}
\end{itemize}

All inference timing analyses are run with a \textit{frame} batch size of one to accurately measure latency and throughput for a realistic streaming input scenario. Note that all pedestrians in a scene at time $t$ are processed simultaneously, and therefore each singular \textit{frame} input still inherently requires a \textit{pedestrian} batch of $P$ trajectories.

\subsection*{Implementation Details}
For the dimensions mentioned in Figure \ref{fig:net}, $\beta$ = 8 and $T$ = 12, in accordance to the evaluation methodology. The MLPs $\phi_0$ and $\phi_1$ contain two hidden layers with input dimensions of 8$\beta$ and 4$\beta$. The embedding layer $\rho$ has an input dimension of $4\beta$ and upscales by 2. We implemented the model in PyTorch\footnote{\url{https://github.com/TeCSAR-UNCC/CARPe_Posterum}} and trained it on an Nvidia Titan V GPU. An open-source PyTorch extension library for graph convolution \cite{geom} was used as the basis for implementing the Graph Module. The model was trained end-to-end with a \textit{frame} batch size of 64 for 80 epochs. We use the Adam \cite{adam} optimizer with a learning rate of 0.01 and a gradient clip of 5. A mean squared error loss was used for training.

\subsection*{Quantitative Results}
\subsubsection{Comparison Approaches:} 
We compare our model to common baseline methods and current state-of-the-art approaches in path prediction. Baseline methods include \textit{Linear}, a simple linear regressor, and Social LSTM \cite{slstm} (\textit{S-LSTM}), a classic method utilizing LSTMs and social pooling. Social GAN \cite{sgan} adds generative models to the Social LSTM approach. \textit{SGAN-P} and \textit{SGAN} indicate the variants with and without the social pooling module as reported in \cite{sgan}. \textit{Sophie} \cite{sophie} employs an LSTM-based GAN module with social and physical attention. Social BiGAT \cite{bigat} (\textit{S-BiGAT}) incorporates LSTMs, Bicycle-GANs \cite{bgan} and physical attention, with GAT \cite{gat} networks to model social elements. \textit{Trajectron} \cite{traj} forms a spatial-temporal graph with a combination of a generative model (CVAEs) and LSTM units. The $z_{best}$ configuration as described in \cite{traj} is used for comparisons. \textit{Next} \cite{next} is a state-of-the-art approach that employs visual pedestrian features and scene segmentation maps for an LSTM-based prediction module with focal attention to make informed trajectory predictions.
\begin{table*}
    \centering
    \renewcommand{\arraystretch}{1.1}
    \begin{adjustbox}{max width=\textwidth}
    \setlength\tabcolsep{4pt}
    \begin{tabular}{l|c|c|c|c|c|c|c|c|c|}
        \textbf{Dataset}    & \textbf{Linear}    & \textbf{S-LSTM}    & \textbf{SGAN-P}    & \textbf{SGAN}      & \textbf{Trajectron} & \textbf{Sophie}         & \textbf{Next}     & \textbf{S-BiGAT }& \textbf{\method}     \\
        \midrule
        \textbf{ETH}        & 1.33 / 2.94        &  1.09 / 2.35       & --                 & 1.13 / 2.21        & 1.13 / 2.44         & --                         & 0.88 / 1.98        & --     & \textbf{0.80} / \textbf{1.48} \\
        \textbf{HOTEL}      & 0.39 / \textbf{0.72} & 0.79 / 1.76      & --                & 1.01 / 2.18        & 0.37 / 0.74          & --                        & \textbf{0.36} / 0.74        & --     & 0.52 / 1.00 \\
        \textbf{UNIV}       & 0.82 / 1.59       & 0.67 / 1.40        & --                 & \textbf{0.60} / 1.28   & 0.62 / 1.31      & --                         & 0.62 / 1.32        & --     & 0.61 / \textbf{1.23} \\
        \textbf{ZARA1}      & 0.62 / 1.21       & 0.47 / 1.00       & --                 & \textbf{0.42} / 0.91    & 0.63 / 1.29      & --                 & \textbf{0.42} / 0.90        & --     & \textbf{0.42} / \textbf{0.84} \\
        \textbf{ZARA2}      & 0.77 / 1.48       & 0.56 / 1.17        & --                 & 0.52 / 1.11    & 0.53 / 1.08             & --                  & \textbf{0.34} / 0.75        & --      & \textbf{0.34} / \textbf{0.69} \\
        \toprule
        \textbf{AVG}        & 0.79 / 1.59      & 0.72 / 1.54      & 0.85 / 1.76*       & 0.74 / 1.54       & 0.66 / 1.40       & 0.71 / 1.46*             & \textbf{0.52} / 1.14           & 0.61 / 1.33*     & 0.54 / \textbf{1.05} \\
        \bottomrule
    \end{tabular}
    \end{adjustbox}
    \caption{ADE and FDE results for all five scenarios in the ETH \cite{eth} and UCY \cite{ucy} datasets. Results followed by * are the K=1 accuracies as reported in the analyzes of \cite{bigat}.}
    \label{tab:quant_res}
\end{table*}

\subsubsection{Accuracy Analysis:} 
ADE and FDE results are reported in Table \ref{tab:quant_res}. It is common for generative approaches in this domain to predict 20 possible trajectories for each pedestrian, and use the closest prediction to ground truth in evaluation. However, since we are considering evaluation within a real-time context, an analysis of the single trajectory prediction results is much more applicable. Therefore, we compare with the K=1 results for all approaches, where K is the number of predictions per pedestrian.

As seen in Table \ref{tab:quant_res}, \method~performs very well against all other methods. In ADE, \method~achieves within 0.02 meters of the best state-of-the-art approach \textit{Next} on average. In FDE, our approach outperforms all other methods. The Prediction Module design to emphasize understanding of velocity change across both the coordinate and social context features allows our method to adjust the final prediction positions accordingly.

Similar to \textit{S-BiGAT}, \method~employs a GNN for gathering insight into the social context. \textit{S-BiGAT} uses GAT operations in its graph, which operate with simple single-layer operations on node features. However, as mathematically shown in \cite{gin}, such single-layer operations are insufficient for robust graph learning. Instead, \method~employs a GIN-based formulation with MLP operations to maximize the discriminative power of its graph module. This design choice in GNN gives \method~a competitive edge over \textit{S-BiGAT}, as revealed in the average ADE and FDE results of Table \ref{tab:quant_res}.

\begin{figure}
	\centering
	\subfloat[Successful examples]{
			\noindent\includegraphics[width=0.95\columnwidth]{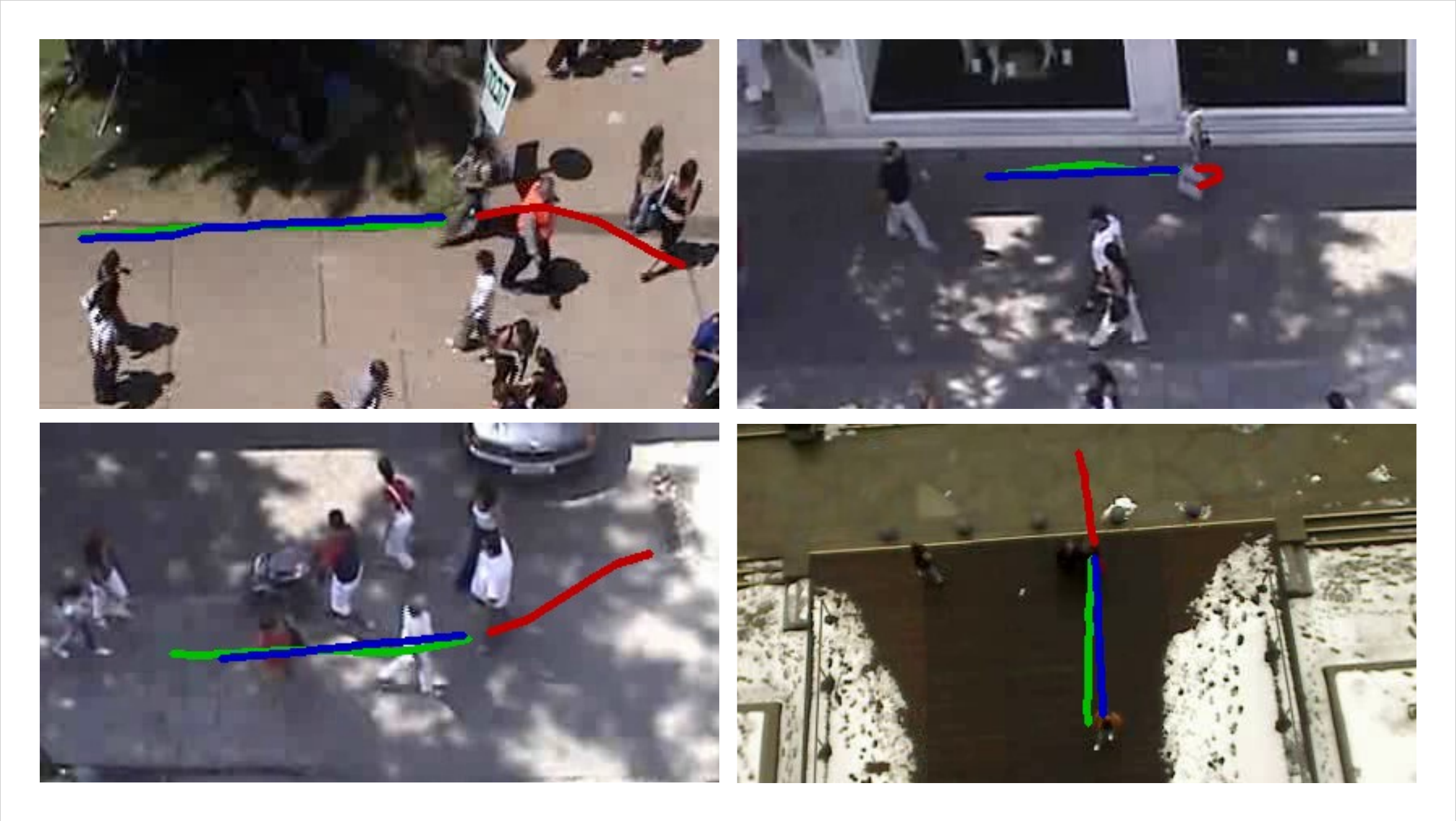}
		}\\
	\subfloat[Failure examples]{
			\noindent\includegraphics[width=0.95\columnwidth]{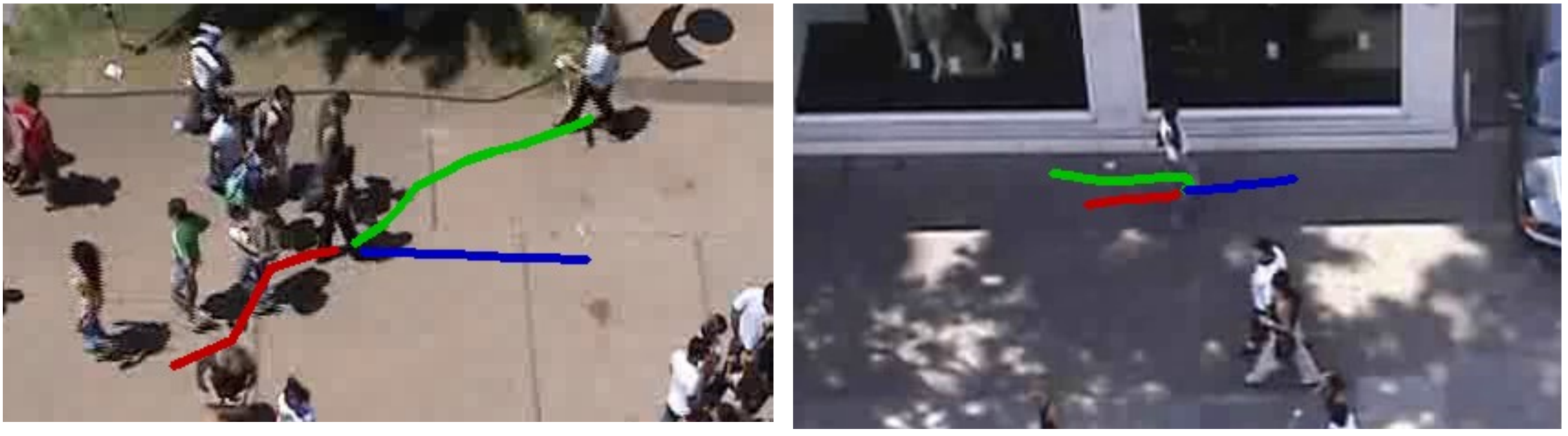}
		}
	\caption{Red indicates the observed trajectory, blue is the predicted trajectory, and green is the ground truth. Images are referenced in to the text as (a.1) to (a.4) and (b.1) to (b.2) from top-left to bottom right.}
	\label{fig:qual}
\end{figure}

\subsubsection{Qualitative Results:}

Figure \ref{fig:qual} illustrates the function of \method~with qualitative examples. The first row of Figure \ref{fig:qual} shows cases of solely intrinsic nonlinearities. In these examples, the pedestrian takes indirect paths with changes in direction and/or speed. In the failure case (b.2), the observed path seems to foreshadow a change in speed alone, but the ground truth indicates that the pedestrian will soon change direction drastically. This abrupt adjustment is not anticipated by \method. However, with some additional information a few time steps later, \method~is able to understand the non-linear behavior and anticipate the future positions of the pedestrian as shown in (a.2). Example (a.1) further illustrates how \method~captures an understanding of speed and direction variations to predict the pedestrian's navigation intent. The second row of Figure \ref{fig:qual} provides samples encompassing intrinsic and social nonlinearities. Example (a.4) shows a situation where the depicted pedestrian alters their trajectory in response to their neighbor's movement as they travel together. Sample (a.3) shows a complex crowd scenario, where \method~is able to determine a likely path through the crowd and achieve a correct prediction. 

The failure case illustrated in (b.2) contains many pedestrians moving individually and in groups. The observed trajectory of the individual traveler is varied in directional intent, and the traveler is nearing collision with a group of pedestrians moving left to right. It is difficult to predict a deterministic route under this intense uncertainty, and therefore \method~takes the safe bet and assumes a path consistent with the social norm (moving left to right with the group). However, the person decides to travel in front of the group and progress upward across the pathway. In situations like this, we note that having visual features of the pedestrians could provide insight into their body position and intended direction. Such modeling would potentially assist in prediction for such scenarios which can be a worthwhile direction for future work.

\begin{table}
    \centering
    \renewcommand{\arraystretch}{1.1}
        \setlength\tabcolsep{4pt}
    \begin{adjustbox}{max width=\columnwidth}
        \begin{tabular}{c|c|c|c|c|c|c|}
            \textbf{Device}   &   \textbf{S-LSTM}    &  \textbf{SGAN-P}   & \textbf{SGAN}  & \textbf{Traj}   & \textbf{Next}    & \textbf{\method}\\ 
            \midrule
            \textbf{P100}   &   0.38                 &  6.67               &     20.00     &    1.04             &    19.46        & \textbf{762.14} \\
            \midrule
        \end{tabular}
        \end{adjustbox}
    \caption{FPS comparison on the Nvidia P100 GPU. Numbers are report as an average per frame across both ETH and UCY datasets.}
    \label{tab:p100}
\end{table}

\subsubsection{Real-time Analysis:}
Path prediction is inherently a time-sensitive task. A crucial characteristic of a path prediction algorithm is its ability to perform real-time inference, particularly on low-power embedded devices. Every fraction of a second is crucial for improving the safety of deployable technologies like self-driving cars and social robots. Therefore, we analyze \method~in comparison to current state-of-the-art approaches for such characteristics. In Table \ref{tab:p100}, we first compare the FPS of \method~on an Nvidia P100 GPU as a baseline. \method~far surpasses the performance of all other methods, by at least 38x across the board.

\method~achieves such improvements for two reasons. First, \method~is designed with real-time inference in mind, eliminating extraneous operations and focused on optimizing the computation expense to accuracy ratio. \\Second, \method~employs a convolutional rather than recurrent architecture. All other methods base their approach on LSTM cells, which limit their hardware utilization capabilities. Instead, \method~is able to take full advantage of the parallel computing capacities of modern hardware, and is thereby well suited for real-world deployment.

In \cite{traj}, \textit{Trajectron} is shown be be significantly faster than SGAN-P when generating 200 samples. However, with N=1, we found that \textit{Trajectron} (Traj in Table \ref{tab:p100}) is slower \textit{SGAN-P}. The inferenced model of \textit{SGAN-P} is faster than that of \textit{Trajectron}, but must be inferened repeatedly to produce each sample, quickly scaling the inference time. \textit{Trajectron} can include additional samples more efficiently; however, the required initial inference of the model is substantially slower due to its graph structure of many recurrent modules.

\textit{Sophie} and \textit{S-BiGAT} do not have their models or latency numbers publicly available, and therefore we do not report their FPS performance. However, it can be noted that both \textit{Sophie} and \textit{S-BiGAT} add additional layers of complexity to the \textit{SGAN} approach. These include the addition of scene-level feature extraction directly on the image using the computationally heavy VGG-19 \cite{vgg19} network, as well as additional scene and social attention mechanisms. Therefore, we can conclude that not only will these approaches experience similar computational difficulties as the other RNN-based architectures, they will also incur additional latencies due to the use of large scene feature extractors and attention networks. 

\begin{table}
    \centering
    \renewcommand{\arraystretch}{1.1}
    \setlength\tabcolsep{4pt}
    \begin{adjustbox}{max width=\columnwidth}
    \begin{tabular}{c|c|c||c|c|}
        \textbf{Approach}    & \textbf{MFLOPs}   & \textbf{Parameters}    & \textbf{FPS (GPU)}    & \textbf{FPS (CPU)}\\
        \midrule
        \textbf{Next}   &    53.08                   & 3.95M              & 5.61               & 5.50\\
        \midrule
        \textbf{\method}   &  \textbf{1.44}        & \textbf{0.10M}         & \textbf{95.87}      & \textbf{48.11}\\
        \midrule
    \end{tabular}
    \end{adjustbox}
    \caption{Detailed comparison with Next \cite{next}. FPS numbers are reported on the Nvidia Jetson Nano embedded device for both GPU and single core CPU. The Nano has a 128-core Maxwell GPU and ARM A57 CPU with a power consumption of approximately 10 Watts in our tests.}
    \label{tab:nano}
\end{table}

In Table \ref{tab:nano}, we thoroughly compare real-time performance of \method~against the best state-of-the-art approach \textit{Next}. First, we analyze the number of floating point operations (FLOPs) and parameters for each approach. \method~is substantially more efficient, reducing the number of FLOPs and parameters by more than 97\%. We also consider the FPS performance on a low-power embedded device, the Nvidia Jetson Nano. For both GPU and single core CPU inference, \method~provides an over 17x and 8x speedup respectively in comparison to \textit{Next}.


Because of its RNN-based design, \textit{Next} is not able to effectively utilize the parallel capabilities of modern hardware, as evident by its almost equal CPU and GPU FPS numbers. In all reported numbers in Tables \ref{tab:p100} and \ref{tab:nano}, we only consider the trajectory generator portion of the \textit{Next} approach. However, this trajectory generator relies on additional scene segmentation features and visual pedestrian features, both of which require large networks for extraction (Xception \cite{xception} and ResNet-101 \cite{resnet} based architectures). On the other hand, \method~successfully employs the GPU architecture, while maintaining fast inference on the CPU. Specifically, with CPU execution, the effect on final inference time for the GNN and CNN is 15\% and 85\% respectively. On the GPU, the GNN and CNN split is 55\% and 45\% respectively. The CNN requires more overall operations, and therefore is responsible for the majority of the CPU time. However, the convolutional operations of the CNN map more efficiently to the GPU hardware than the aggregate/combine operations of the GNN. Therefore, the effect on inference time between the GNN and CNN becomes relatively close on the GPU.



\section*{Conclusions}
We proposed \method, a convolutional approach for real-time pedestrian path prediction. Distinct from prior work, \method~is able to produce accurate future trajectory predictions within real-time constraints. \method~is an agile CNN that operates across the temporal context of observations in the spatial domain, maximizing both feature correlation and parallel hardware utilization. We also employed a discriminative graph neural network based on GIN operators to gather social context features, providing additional insight into the predictive model. \method~captures non-linear intrinsic and social effects, achieving competitive accuracy results in comparison with the current SotA methods while enabling 8x to 38x improvements in FPS. 

\section*{ Acknowledgments}
This research is supported by the National Science Foundation (NSF) under Awards No. 1831795.

\bibliography{main}

\begin{thebibliography}{42}
\providecommand{\natexlab}[1]{#1}
\providecommand{\url}[1]{\texttt{#1}}
\providecommand{\urlprefix}{URL }
\expandafter\ifx\csname urlstyle\endcsname\relax
  \providecommand{\doi}[1]{doi:\discretionary{}{}{}#1}\else
  \providecommand{\doi}{doi:\discretionary{}{}{}\begingroup
  \urlstyle{rm}\Url}\fi

\bibitem[{Alahi et~al.(2016)Alahi, Goel, Ramanathan, Robicquet, Fei-Fei, and
  Savarese}]{slstm}
Alahi, A.; Goel, K.; Ramanathan, V.; Robicquet, A.; Fei-Fei, L.; and Savarese,
  S. 2016.
\newblock Social LSTM: Human Trajectory Prediction in Crowded Spaces.
\newblock In \emph{The IEEE Conference on Computer Vision and Pattern
  Recognition (CVPR)}.

\bibitem[{Bai, Kolter, and Koltun(2018)}]{cnn2}
Bai, S.; Kolter, J.~Z.; and Koltun, V. 2018.
\newblock An Empirical Evaluation of Generic Convolutional and Recurrent
  Networks for Sequence Modeling.
\newblock \emph{CoRR} abs/1803.01271.
\newblock \urlprefix\url{http://arxiv.org/abs/1803.01271}.

\bibitem[{Chollet(2016)}]{xception}
Chollet, F. 2016.
\newblock Xception: Deep Learning with Depthwise Separable Convolutions.
\newblock \emph{CoRR} abs/1610.02357.
\newblock \urlprefix\url{http://arxiv.org/abs/1610.02357}.

\bibitem[{Defferrard, Bresson, and Vandergheynst(2016)}]{cheb}
Defferrard, M.; Bresson, X.; and Vandergheynst, P. 2016.
\newblock Convolutional neural networks on graphs with fast localized spectral
  filtering.
\newblock In \emph{Proceedings of the 30th International Conference on Neural
  Information Processing Systems}, 3844--3852.

\bibitem[{Elbayad, Besacier, and Verbeek(2018)}]{cnn1}
Elbayad, M.; Besacier, L.; and Verbeek, J. 2018.
\newblock Pervasive Attention: 2D Convolutional Neural Networks for
  Sequence-to-Sequence Prediction.

\bibitem[{Fey and Lenssen(2019)}]{geom}
Fey, M.; and Lenssen, J.~E. 2019.
\newblock Fast Graph Representation Learning with {PyTorch Geometric}.
\newblock In \emph{ICLR Workshop on Representation Learning on Graphs and
  Manifolds}.

\bibitem[{Gehring et~al.(2017)Gehring, Auli, Grangier, Yarats, and
  Dauphin}]{convseq}
Gehring, J.; Auli, M.; Grangier, D.; Yarats, D.; and Dauphin, Y.~N. 2017.
\newblock Convolutional sequence to sequence learning.
\newblock \emph{arXiv preprint arXiv:1705.03122} .

\bibitem[{Goodfellow, Bengio, and Courville(2016)}]{goodfellow}
Goodfellow, I.; Bengio, Y.; and Courville, A. 2016.
\newblock \emph{Deep Learning}.
\newblock MIT Press.
\newblock \url{http://www.deeplearningbook.org}.

\bibitem[{Goodfellow et~al.(2014)Goodfellow, Pouget-Abadie, Mirza, Xu,
  Warde-Farley, Ozair, Courville, and Bengio}]{gan}
Goodfellow, I.; Pouget-Abadie, J.; Mirza, M.; Xu, B.; Warde-Farley, D.; Ozair,
  S.; Courville, A.; and Bengio, Y. 2014.
\newblock Generative Adversarial Nets.
\newblock In Ghahramani, Z.; Welling, M.; Cortes, C.; Lawrence, N.~D.; and
  Weinberger, K.~Q., eds., \emph{Advances in Neural Information Processing
  Systems 27}, 2672--2680. Curran Associates, Inc.
\newblock
  \urlprefix\url{http://papers.nips.cc/paper/5423-generative-adversarial-nets.pdf}.

\bibitem[{Graves(2013)}]{rnn_seq}
Graves, A. 2013.
\newblock Generating Sequences With Recurrent Neural Networks.
\newblock \emph{CoRR} abs/1308.0850.
\newblock \urlprefix\url{http://arxiv.org/abs/1308.0850}.

\bibitem[{{Gupta} et~al.(2018){Gupta}, {Johnson}, {Fei-Fei}, {Savarese}, and
  {Alahi}}]{sgan}
{Gupta}, A.; {Johnson}, J.; {Fei-Fei}, L.; {Savarese}, S.; and {Alahi}, A.
  2018.
\newblock Social GAN: Socially Acceptable Trajectories with Generative
  Adversarial Networks.
\newblock In \emph{2018 IEEE/CVF Conference on Computer Vision and Pattern
  Recognition}, 2255--2264.
\newblock ISSN 1063-6919.
\newblock \doi{10.1109/CVPR.2018.00240}.

\bibitem[{Hamilton, Ying, and Leskovec(2017)}]{sage}
Hamilton, W.; Ying, Z.; and Leskovec, J. 2017.
\newblock Inductive representation learning on large graphs.
\newblock In \emph{Advances in neural information processing systems},
  1024--1034.

\bibitem[{He et~al.(2015)He, Zhang, Ren, and Sun}]{resnet}
He, K.; Zhang, X.; Ren, S.; and Sun, J. 2015.
\newblock Deep Residual Learning for Image Recognition.
\newblock \emph{CoRR} abs/1512.03385.
\newblock \urlprefix\url{http://arxiv.org/abs/1512.03385}.

\bibitem[{Helbing and Molnár(1995)}]{social_force}
Helbing, D.; and Molnár, P. 1995.
\newblock Social force model for pedestrian dynamics.
\newblock \emph{Physical Review E} 51(5): 4282–4286.
\newblock ISSN 1095-3787.
\newblock \doi{10.1103/physreve.51.4282}.
\newblock \urlprefix\url{http://dx.doi.org/10.1103/PhysRevE.51.4282}.

\bibitem[{Hochreiter and Schmidhuber(1997)}]{lstm}
Hochreiter, S.; and Schmidhuber, J. 1997.
\newblock Long short-term memory.
\newblock \emph{Neural computation} 9(8): 1735--1780.

\bibitem[{Hornik(1991)}]{uat2}
Hornik, K. 1991.
\newblock Approximation capabilities of multilayer feedforward networks.
\newblock \emph{Neural networks} 4(2): 251--257.

\bibitem[{Hornik et~al.(1989)Hornik, Stinchcombe, White et~al.}]{uat1}
Hornik, K.; Stinchcombe, M.; White, H.; et~al. 1989.
\newblock Multilayer feedforward networks are universal approximators.
\newblock \emph{Neural networks} 2(5): 359--366.

\bibitem[{Ivanovic and Pavone(2019)}]{traj}
Ivanovic, B.; and Pavone, M. 2019.
\newblock The Trajectron: Probabilistic Multi-Agent Trajectory Modeling With
  Dynamic Spatiotemporal Graphs.
\newblock In \emph{Proceedings of the IEEE/CVF International Conference on
  Computer Vision (ICCV)}.

\bibitem[{Kingma and Ba(2014)}]{adam}
Kingma, D.~P.; and Ba, J. 2014.
\newblock Adam: A method for stochastic optimization.
\newblock \emph{arXiv preprint arXiv:1412.6980} .

\bibitem[{Kipf and Welling(2016)}]{GCN}
Kipf, T.~N.; and Welling, M. 2016.
\newblock Semi-supervised classification with graph convolutional networks.
\newblock \emph{arXiv preprint arXiv:1609.02907} .

\bibitem[{Kitani et~al.(2012)Kitani, Ziebart, Bagnell, and
  Hebert}]{irl_activity}
Kitani, K.~M.; Ziebart, B.~D.; Bagnell, J.~A.; and Hebert, M. 2012.
\newblock Activity Forecasting.
\newblock In Fitzgibbon, A.; Lazebnik, S.; Perona, P.; Sato, Y.; and Schmid,
  C., eds., \emph{Computer Vision -- ECCV 2012}, 201--214. Berlin, Heidelberg:
  Springer Berlin Heidelberg.
\newblock ISBN 978-3-642-33765-9.

\bibitem[{Kosaraju et~al.(2019)Kosaraju, Sadeghian,
  Mart{\'{\i}}n{-}Mart{\'{\i}}n, Reid, Rezatofighi, and Savarese}]{bigat}
Kosaraju, V.; Sadeghian, A.; Mart{\'{\i}}n{-}Mart{\'{\i}}n, R.; Reid, I.~D.;
  Rezatofighi, S.~H.; and Savarese, S. 2019.
\newblock Social-BiGAT: Multimodal Trajectory Forecasting using Bicycle-GAN and
  Graph Attention Networks.
\newblock \emph{CoRR} abs/1907.03395.
\newblock \urlprefix\url{http://arxiv.org/abs/1907.03395}.

\bibitem[{Lerner, Chrysanthou, and Lischinski(2007)}]{ucy}
Lerner, A.; Chrysanthou, Y.; and Lischinski, D. 2007.
\newblock Crowds by Example.
\newblock \emph{Comput. Graph. Forum} 26: 655--664.

\bibitem[{Li(2019)}]{idl}
Li, Y. 2019.
\newblock Which Way Are You Going? Imitative Decision Learning for Path
  Forecasting in Dynamic Scenes.
\newblock In \emph{The IEEE Conference on Computer Vision and Pattern
  Recognition (CVPR)}.

\bibitem[{Liang et~al.(2019)Liang, Jiang, Niebles, Hauptmann, and
  Fei-Fei}]{next}
Liang, J.; Jiang, L.; Niebles, J.~C.; Hauptmann, A.~G.; and Fei-Fei, L. 2019.
\newblock Peeking into the future: Predicting future person activities and
  locations in videos.
\newblock In \emph{Proceedings of the IEEE Conference on Computer Vision and
  Pattern Recognition}, 5725--5734.

\bibitem[{Manh and Alaghband(2018)}]{scene-lstm}
Manh, H.; and Alaghband, G. 2018.
\newblock Scene-LSTM: {A} Model for Human Trajectory Prediction.
\newblock \emph{CoRR} abs/1808.04018.
\newblock \urlprefix\url{http://arxiv.org/abs/1808.04018}.

\bibitem[{{Pellegrini} et~al.(2009{\natexlab{a}}){Pellegrini}, {Ess},
  {Schindler}, and {van Gool}}]{youll_never_walk_alone}
{Pellegrini}, S.; {Ess}, A.; {Schindler}, K.; and {van Gool}, L.
  2009{\natexlab{a}}.
\newblock You'll never walk alone: Modeling social behavior for multi-target
  tracking.
\newblock In \emph{2009 IEEE 12th International Conference on Computer Vision},
  261--268.
\newblock ISSN 1550-5499.
\newblock \doi{10.1109/ICCV.2009.5459260}.

\bibitem[{{Pellegrini} et~al.(2009{\natexlab{b}}){Pellegrini}, {Ess},
  {Schindler}, and {van Gool}}]{eth}
{Pellegrini}, S.; {Ess}, A.; {Schindler}, K.; and {van Gool}, L.
  2009{\natexlab{b}}.
\newblock You'll never walk alone: Modeling social behavior for multi-target
  tracking.
\newblock In \emph{2009 IEEE 12th International Conference on Computer Vision},
  261--268.
\newblock ISSN 1550-5499.
\newblock \doi{10.1109/ICCV.2009.5459260}.

\bibitem[{Pellegrini, Ess, and Van~Gool(2010)}]{joint_modeling}
Pellegrini, S.; Ess, A.; and Van~Gool, L. 2010.
\newblock Improving Data Association by Joint Modeling of Pedestrian
  Trajectories and Groupings.
\newblock In Daniilidis, K.; Maragos, P.; and Paragios, N., eds.,
  \emph{Computer Vision -- ECCV 2010}, 452--465. Berlin, Heidelberg: Springer
  Berlin Heidelberg.
\newblock ISBN 978-3-642-15549-9.

\bibitem[{Rudenko et~al.(2019)Rudenko, Palmieri, Herman, Kitani, Gavrila, and
  Arras}]{path_survey}
Rudenko, A.; Palmieri, L.; Herman, M.; Kitani, K.~M.; Gavrila, D.~M.; and
  Arras, K.~O. 2019.
\newblock Human Motion Trajectory Prediction: {A} Survey.
\newblock \emph{CoRR} abs/1905.06113.
\newblock \urlprefix\url{http://arxiv.org/abs/1905.06113}.

\bibitem[{Sadeghian et~al.(2019)Sadeghian, Kosaraju, Sadeghian, Hirose,
  Rezatofighi, and Savarese}]{sophie}
Sadeghian, A.; Kosaraju, V.; Sadeghian, A.; Hirose, N.; Rezatofighi, H.; and
  Savarese, S. 2019.
\newblock SoPhie: An Attentive GAN for Predicting Paths Compliant to Social and
  Physical Constraints.
\newblock In \emph{The IEEE Conference on Computer Vision and Pattern
  Recognition (CVPR)}.

\bibitem[{Simonyan and Zisserman(2014)}]{vgg19}
Simonyan, K.; and Zisserman, A. 2014.
\newblock Very deep convolutional networks for large-scale image recognition.
\newblock \emph{arXiv preprint arXiv:1409.1556} .

\bibitem[{Sohn, Lee, and Yan(2015)}]{cvae}
Sohn, K.; Lee, H.; and Yan, X. 2015.
\newblock Learning Structured Output Representation using Deep Conditional
  Generative Models.
\newblock In Cortes, C.; Lawrence, N.~D.; Lee, D.~D.; Sugiyama, M.; and
  Garnett, R., eds., \emph{Advances in Neural Information Processing Systems
  28}, 3483--3491. Curran Associates, Inc.
\newblock
  \urlprefix\url{http://papers.nips.cc/paper/5775-learning-structured-output-representation-using-deep-conditional-generative-models.pdf}.

\bibitem[{Tay and Laugier(2008)}]{smooth_guassian}
Tay, M. K.~C.; and Laugier, C. 2008.
\newblock \emph{Modelling Smooth Paths Using Gaussian Processes}, 381--390.
\newblock Berlin, Heidelberg: Springer Berlin Heidelberg.
\newblock ISBN 978-3-540-75404-6.
\newblock \doi{10.1007/978-3-540-75404-6_36}.

\bibitem[{Veličković et~al.(2018)Veličković, Cucurull, Casanova, Romero,
  Liò, and Bengio}]{gat}
Veličković, P.; Cucurull, G.; Casanova, A.; Romero, A.; Liò, P.; and Bengio,
  Y. 2018.
\newblock Graph Attention Networks.
\newblock In \emph{International Conference on Learning Representations}.
\newblock \urlprefix\url{https://openreview.net/forum?id=rJXMpikCZ}.

\bibitem[{Weisfeiler and Lehman(1968)}]{wl}
Weisfeiler, B.; and Lehman, A.~A. 1968.
\newblock A reduction of a graph to a canonical form and an algebra arising
  during this reduction.
\newblock \emph{Nauchno-Technicheskaya Informatsia} 2(9): 12--16.

\bibitem[{{Wu} et~al.(2021){Wu}, {Pan}, {Chen}, {Long}, {Zhang}, and
  {Yu}}]{survey}
{Wu}, Z.; {Pan}, S.; {Chen}, F.; {Long}, G.; {Zhang}, C.; and {Yu}, P.~S. 2021.
\newblock A Comprehensive Survey on Graph Neural Networks.
\newblock \emph{IEEE Transactions on Neural Networks and Learning Systems}
  32(1): 4--24.
\newblock \doi{10.1109/TNNLS.2020.2978386}.

\bibitem[{Xu et~al.(2019)Xu, Hu, Leskovec, and Jegelka}]{gin}
Xu, K.; Hu, W.; Leskovec, J.; and Jegelka, S. 2019.
\newblock How Powerful are Graph Neural Networks?
\newblock In \emph{International Conference on Learning Representations}.
\newblock \urlprefix\url{https://openreview.net/forum?id=ryGs6iA5Km}.

\bibitem[{{Xu}, {Piao}, and {Gao}(2018)}]{cidnn}
{Xu}, Y.; {Piao}, Z.; and {Gao}, S. 2018.
\newblock Encoding Crowd Interaction with Deep Neural Network for Pedestrian
  Trajectory Prediction.
\newblock In \emph{2018 IEEE/CVF Conference on Computer Vision and Pattern
  Recognition}, 5275--5284.
\newblock \doi{10.1109/CVPR.2018.00553}.

\bibitem[{{Yamaguchi} et~al.(2011){Yamaguchi}, {Berg}, {Ortiz}, and
  {Berg}}]{who_are_you_with}
{Yamaguchi}, K.; {Berg}, A.~C.; {Ortiz}, L.~E.; and {Berg}, T.~L. 2011.
\newblock Who are you with and where are you going?
\newblock In \emph{CVPR 2011}, 1345--1352.
\newblock ISSN 1063-6919.
\newblock \doi{10.1109/CVPR.2011.5995468}.

\bibitem[{Zhang et~al.(2019)Zhang, Ouyang, Zhang, Xue, and Zheng}]{srlstm}
Zhang, P.; Ouyang, W.; Zhang, P.; Xue, J.; and Zheng, N. 2019.
\newblock Sr-lstm: State refinement for lstm towards pedestrian trajectory
  prediction.
\newblock In \emph{Proceedings of the IEEE Conference on Computer Vision and
  Pattern Recognition}, 12085--12094.

\bibitem[{Zhu et~al.(2017)Zhu, Zhang, Pathak, Darrell, Efros, Wang, and
  Shechtman}]{bgan}
Zhu, J.-Y.; Zhang, R.; Pathak, D.; Darrell, T.; Efros, A.~A.; Wang, O.; and
  Shechtman, E. 2017.
\newblock Toward multimodal image-to-image translation.
\newblock In \emph{Advances in neural information processing systems},
  465--476.

\end{thebibliography}

\end{document}